\documentclass{article}
\usepackage{microtype}
\usepackage{graphicx}
\usepackage{subfigure}
\usepackage{booktabs} 
\usepackage{physics}
\usepackage{caption}
\usepackage{hyperref}

\usepackage[accepted]{icml2025}

\usepackage{amsmath}
\usepackage{amssymb}
\usepackage{mathtools}
\usepackage{amsthm}
\usepackage{makecell}
\usepackage[capitalize,noabbrev]{cleveref}

\theoremstyle{plain}

\theoremstyle{definition}

\theoremstyle{remark}

\usepackage[textsize=tiny]{todonotes}

\icmltitlerunning{Frankenstein Optimizer}

\begin{document}

\twocolumn[
    \icmltitle{Frankenstein Optimizer: Harnessing the Potential by Revisiting Optimization Tricks}
    
    \icmlsetsymbol{equal}{*}
    
    \begin{icmlauthorlist}
    \icmlauthor{Chia-Wei Hsu}{DMSE}
    \icmlauthor{Nien-Ti Tsou}{NYCU}
    \icmlauthor{Yu-Cheng Chen}{HH}
    \icmlauthor{Yang Jeong Park}{DNSE}
    \icmlauthor{Ju Li}{DMSE,DNSE}
    \end{icmlauthorlist}
    
    \icmlaffiliation{DMSE}{Department of Materials Science and Engineering, Massachusetts Institute of Technology, Cambridge, MA 02139}
    \icmlaffiliation{DNSE}{Department of Nuclear Science and Engineering, Massachusetts Institute of Technology, Cambridge, MA 02139}
    \icmlaffiliation{NYCU}{Department of Materials Science and Engineering, National Yang Ming Chiao Tung University, Taiwan}
    \icmlaffiliation{HH}{Hon Hai Research Institute, Taipei, Taiwan}
    
    \icmlcorrespondingauthor{Yang Jeong Park}{parkyj@mit.edu}
    \icmlcorrespondingauthor{Ju Li}{liju@mit.edu}

    \icmlkeywords{Deep Learning, Optimization, ICML}
    \printAffiliationsAndNotice{} 
    \vskip 0.3in
]




\begin{abstract}
Gradient-based optimization drives the unprecedented performance of modern deep neural network models across diverse applications. Adaptive algorithms have accelerated neural network training due to their rapid convergence rates; however, they struggle to find ``flat minima" reliably, resulting in suboptimal generalization compared to stochastic gradient descent (SGD). By revisiting various adaptive algorithms' mechanisms, we propose the Frankenstein optimizer, which combines their advantages. The proposed Frankenstein dynamically adjusts first- and second-momentum coefficients according to the optimizer's current state to directly maintain consistent learning dynamics and immediately reflect sudden gradient changes. Extensive experiments across several research domains such as computer vision, natural language processing, few-shot learning, and scientific simulations show that Frankenstein surpasses existing adaptive algorithms and SGD empirically regarding convergence speed and generalization performance. Furthermore, this research deepens our understanding of adaptive algorithms through centered kernel alignment analysis and loss landscape visualization during the learning process. Code is available at \url{https://github.com/acctouhou/Frankenstein_optimizer}
\end{abstract}

\section{Introduction}
In recent years, neural networks (NNs) have become increasingly prevalent in various fields, driving the widespread adoption of gradient-based optimization methods in science and engineering. This increased usage has led to studies enhancing optimizers to achieve faster convergence and better results with limited computational resources. Among these optimizers, Adam\cite{adam} has emerged as the most prominent, often serving as the default setting in deep learning tasks. However, many studies have highlighted that Adam-like optimizers do not necessarily achieve superior generalization\cite{Margin1,Margin2}, with simple experiments\cite{simple_test} demonstrating this issue. As a result, the debate between Adam and stochastic gradient descent (SGD) has persisted within the deep learning community.

The rapid growth in the parameter size of deep learning models has magnified the importance of adaptive optimizers despite their known limitations \cite{parameter_size}. Current state-of-the-art large language models (LLMs), such as Nemotron-4 with 340 billion parameters \cite{340b}, LLaMA 3.1 with 405 billion parameters\cite{llama31} and DeepSeek-V3 with 671 billion parameters \cite{deepseek}, exemplify the growing scale of modern architectures. Considering the computational cost for training GPT-3 with 175 billion parameters \cite{gpt3} which amounts to \$4.6 million, it is obvious that the efficiency of optimization algorithms directly impacts AI economics.

Adam or its variants have been actively developed to design faster optimizers. For example, AdaBelief\cite{adabelief} demonstrates its superiority in addressing deep valley problems by simultaneously considering both adaptive terms and momentum. Similarly, Sophia optimizer\cite{sophia} showcases its ability to manage risks in the update process by correcting the Hessian, ensuring its stability, and yielding promising results in large models like LLMs. However, the adaptive nature during optimization remains underexplored, leaving room for further improvement.

\begin{figure}[ht]
    \centering
    \includegraphics[width=\linewidth]{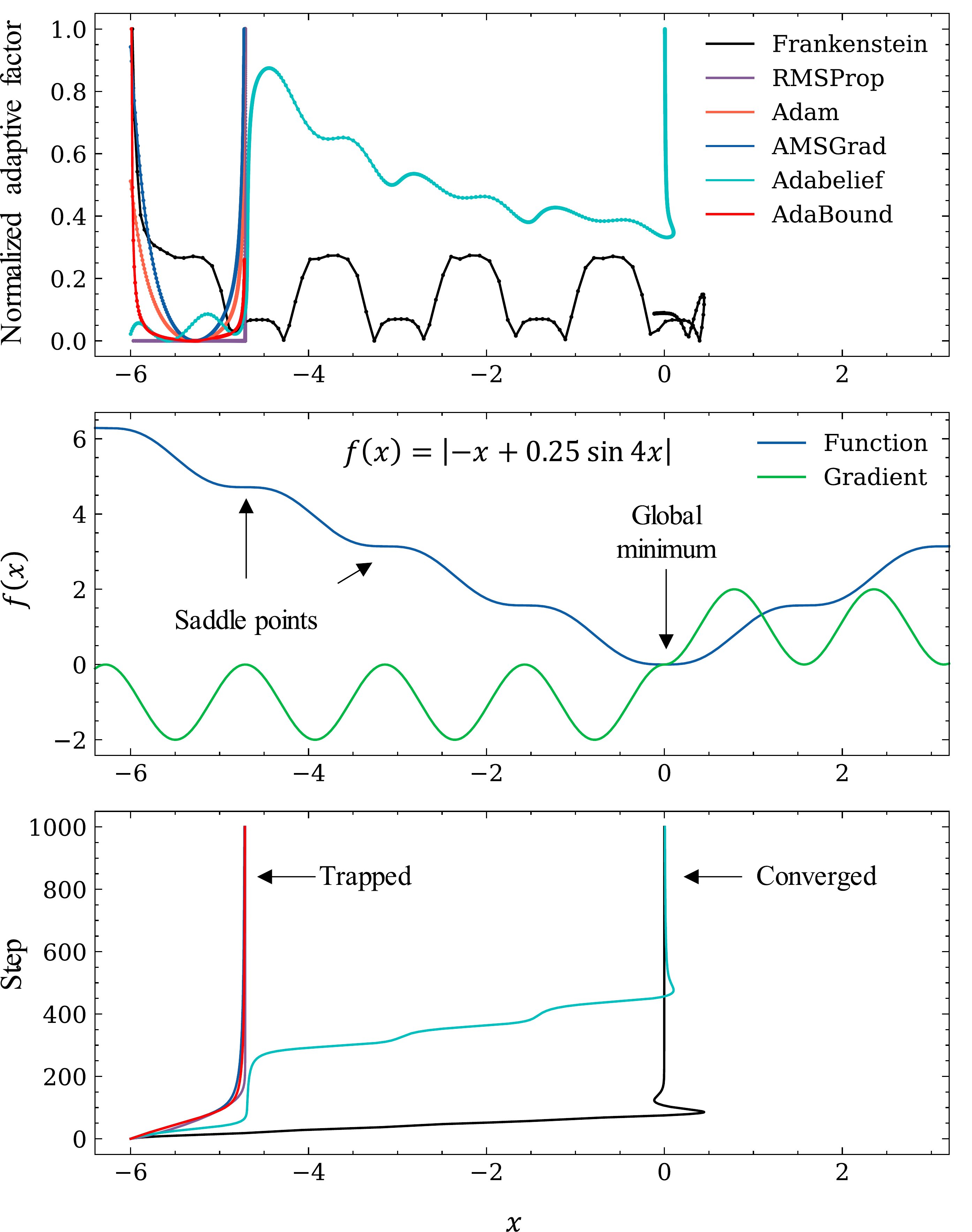}
    \caption{From top to bottom: (1) Adaptive factor for each optimizer during the optimization process of \emph{f}(x); (2) Function \emph{f}(x) and its corresponding gradient; (3) Historical trajectories of each optimizer as a function of optimization steps.}
    \label{fig:demo}
\end{figure}

To further explore these adaptive adjustments and the potential for improvement, we propose a new optimizer, ``Frankenstein,'' developed on the techniques used by previous gradient-based adaptive optimizers, while incorporating our three key innovations: 
\begin{itemize}
\item \textbf{1) Adaptive momentum coefficients $\beta$}: Unlike previous optimizers that employ a fixed momentum coefficient, our approach dynamically adjusts the momentum based on the current state and learning rate schedule. This design enhances the adaptability of the optimization process, enabling more efficient convergence across diverse training scenarios.
\item \textbf{2) Relaxation strategy for second momentum $v_{t}$ }: Along with adjusting the second momentum, we propose a novel relaxation strategy that balances the maximum recorded value, as utilized in AMSGrad variants \cite{amsgrad}, with conventional unconstrained exponential moving average methods. This approach accelerates optimization while mitigating convergence risks, resulting in significant improvements in solving adaptive problems.
\item \textbf{3) Acceleration factor $\xi$}: For fast and accurate convergence, we designed an acceleration factor that comprehensively evaluates the system's gradient magnitude and degree of belief, as well as its proximity to convergence. It determines how much the gradient is reflected in the parameter update. 
\end{itemize}
To briefly demonstrate the effect of those design principles, we visualize optimization process of various existing adaptive algorithms and Frankenstein on a loss function, as shown in Fig.\ref{fig:demo}. In scenarios with larger gradients, our method achieves acceleration up to 1.6 times the standard learning rate. However, as convergence approaches, updates rely primarily on the second momentum term, which decays more rapidly, making the nonlinear misalignment factor $P$ the primary factor driving acceleration. This produces two distinct peaks in the normalized adaptive factor. The adaptive factor is defined as any value applied to momentum or gradients during the position update that controls the learning rate. The adaptive factor throughout the entire learning process contributes to understanding the optimizer's preferences and behavior. This switching mechanism also achieves robust convergence in plateaued landscapes\cite{plateaus}, which is a significant issue for quantum NN training. $\xi$ will ultimately revert to a value of 1, which is the non-accelerated state, due to the vanishing gradient and the convergence of nonlinear $P$ factor.

In the remainder of this paper, we provide a more detailed explanation of the algorithm \ref{alg:frank_opt} and validate its superiority through experiments across various applications. Various visualization analyses during the learning process were conducted to gain a deeper understanding of the algorithm's adaptability.

\section{Algorithm}\label{sec:algorithm}
\subsection{Adaptive first momentum coefficient $\beta$}

Adjusting the learning rate throughout NN training is a common strategy to improve model performance. Research indicates\cite{decaylearningrate,learn_lr,Sharp_minima2} that there should be a consistent ``noise scale'' across batch size, learning rate, and momentum. Therefore, our proposed algorithm also adapts the momentum coefficient based on the current learning rate. In Verlet integration, $\Delta x$ is affected by the current momentum, gradient, and learning rate, as illustrated in Equation~\ref{eq:verlet}. Let $\alpha$ represent the learning rate, $m_t$ represent the first-order momentum, $g_t$ represent the gradient, and $\theta$ represent the learning parameter.
\begin{equation}
    \theta_{t+1} = \theta_t + m_t \alpha + \frac{1}{2} g_t \alpha^2
    \label{eq:verlet}
\end{equation}
If the training process enters a stable phase where the gradient remains relatively constant and the momentum closely matches the gradient, the momentum coefficient should be maintained as specified in Equation~\ref{eq:tie}; otherwise, this balance could be disrupted.
\begin{equation}
    (1 - \beta) m_t \approx \frac{1}{2} g_t \alpha^2
    \label{eq:tie}
\end{equation}

To maintain training consistency, the decay rate caused by momentum and the momentum coefficient should correspond to the gradient and the learning rate; otherwise, using a learning rate decay strategy might lead to overly rapid convergence, requiring additional iterations to achieve the optimal convergence point. Based on this insight, we propose a momentum coefficient adjustment strategy, detailed in Equation~\ref{eq:momentum_adjustment}. This formula reflects common configurations, where training from scratch uses $\alpha = 10^{-3}$ and $\beta = 0.9$, while fine-tuning adopts $\alpha = 2 \times 10^{-4}$ and $\beta = 0.9$. In our algorithm, each time the learning rate changes, a new optimal $\beta^*$ is calculated based on the corresponding learning rate instead of using a fixed $\beta$.

\begin{equation}
     \beta^* = 1 - (1 - \beta) \sqrt{\frac{ \alpha}{10^{-3}}}
     \label{eq:momentum_adjustment}
 \end{equation}

It’s worth noting that other algorithms update momentum by applying a fixed-coefficient moving average, which retains information from multiple previous steps during iterations and smooths the trajectory. This approach takes several steps to approximate the ideal state. In contrast, our Frankenstein can directly maintain consistent learning dynamics. In other words, it isn’t forced to converge due to changes in the learning rate; instead, it decides whether to converge based on the actual gradient conditions (with the final momentum coefficient still remaining $< 1$).

Here, we define the nonlinear misalignment factor ($P$), which indicates a misalignment between the past momentum $m_{t-1}$ and current gradient $g_t$. $P$ takes a value between 0 and 1, and it becomes smaller when the two vectors are well-aligned in the same direction, while it becomes larger when they are aligned in opposite directions. 
\begin{equation}
    P \gets \cos^{-1} (\tanh(m_{t-1} \odot g_t)) / \pi
\end{equation}
Additionally, the we propose another parameter $\rho$ to adjust the momentum coefficient. 
\begin{equation}
    \rho \gets \log(e^{1} + \sqrt{x_t} + 0.5 - P)
\end{equation}
where $x_t$ refers the summation of squared gradient $g_t^2$ and a small constant $\epsilon$. The $\rho$ parameter controls whether the momentum coefficient should increase based on the difference between the current system state and the ideal convergence state. The coefficient is influenced by two conditions: (1) whether the squared gradient approaches zero and (2) whether the $P$ factor approaches 0.5. As these conditions near the convergence state, the momentum coefficient will gradually decrease to the default value of $\beta$, effectively rendering it inactive.

Notably, the momentum coefficient can be adjusted to zero if the product of the momentum direction and the gradient is highly negative. This mechanism is inspired by the FIRE algorithm\cite{fire_lammps}, where a large directional discrepancy may indicate that the step size is too large, causing a lack of convergence. Therefore, this feature includes reinitializing the momentum or reversing its direction when necessary.

\subsection{Second momentum $v_{t}$ update with dynamic EMA}

The update rules for second momentum $v_t$ enable various strategies to achieve optimizers' adaptivity. The general process is outlined in the formula below.
\begin{align}
g_t & \leftarrow \nabla_\theta f_t (\theta_{t-1}) \\
v_t & \leftarrow V(v_{t-1},g_t, \beta,...) \\
\theta_{t+1} & \leftarrow U(\theta_{t}, v_t, \alpha,...)
\label{eq:generalized_formula}
\end{align}
where $V$ represents a momentum update function and $U$ represents a parameter update function.

Starting with the basic RMSProp strategy, which applies a simple Exponential Moving Average (EMA) to the squared gradients, several adaptations have been developed (Appendix \ref{sec:existing_optimizers}). These methods depends on a fixed $\beta_2$ parameter throughout the training process. In contrast, this work proposes a strategy that also relies on squared gradients as its foundation but applies an EMA coefficient that adapts based on both current and historical states. This coefficient is determined by whether the $P$ factor approaches 0.5 and the ratio of the current to past squared gradients. The following paragraphs explain the motivations for this adjustment.

These Adam-like optimizers employ adaptive methods with second-order momentum, using a higher momentum coefficient for EMA compared to first-order momentum. This setup not only facilitates the transition of the Adam phase from divergence to convergence\cite{adam_beta_converge,amsgrad} but also allows a larger $\beta_2$ to enhance long-term memory, smoothing the training process and helping to prevent issues like gradient instability. However, as NN parameters scale up, LLMs now use a $\beta_2$ value of 0.95 instead of the previous 0.999. 

Even when addressing instability in the training process of large models by tuning $\beta_2$, potential spikes in the training trajectory may still occur, where a fixed $\beta_2$ can cause the adaptive ratio,
\begin{equation}
    r_t = \frac{m_t}{\sqrt{v_t}}
\end{equation}
to shift from a unimodal distribution approaching zero gradient to a bimodal one, which may destabilize LLM training. In other words, fixing $\beta_2$ establishes a range within which gradient fluctuations remain manageable, especially avoiding sharp increases. Otherwise, a slow EMA could introduce bias in the adaptive coefficient, ultimately harming model performance.\cite{adam_llm,spike_loss}

This study addresses the challenge associated with the $\beta_2$ coefficient by introducing a dynamic $\beta_2$ that adjusts according to the current training state, enhancing adaptability and stability. The approach leverages the ratio between the current and past squared gradients, allowing sudden gradients to be immediately reflected in $v_t$ and strengthening the convergence stability. 
\begin{equation}
    \beta_{2} \gets 1-\frac{x_t}{x_{t-1}} |0.5 - P|
\end{equation}
Notably, in cases where the current-to-past squared gradient ratio exceeds 1, $\beta_2$ may even take on negative values, further enhancing adaptability during drastic gradient changes. This mechanism outperforms a simple $\beta_2$ setting of zero by providing better adaptive control, as shown in the equation below.
\begin{equation}
v_t = g_t^2 - \beta_2(g_t^2 - v_{t-1})
\end{equation}
\begin{equation}
\text{where } \beta_2(g_t^2 - v_{t-1}) > 0
\end{equation}



Another critical aspect of our $\beta_2$ design is including the $P$ factor, which also serves as an indicator of convergence. When the $P$ factor is near 0, it reveals whether the directions of the gradient and momentum are aligned and provides insight into their relative magnitudes. This makes it a reliable measure of how close the training process is to convergence. Moreover, the reduced $\beta_2$ allows the current gradient to contribute more significantly to the second momentum update, giving the algorithm greater ``confidence'' in its convergence.  

When gradients begin to diminish—a phenomenon known as gradient vanishing—the influence of the $P$ factor allows $\beta_2$ to approach 1, effectively maintaining long-term memory. This adjustment offers two significant benefits: it enables the model to better handle high-plateau regions, where gradients tend to be low, and it encourages convergence toward flatter minima. Such convergence is beneficial for achieving smoother and potentially more generalizable solutions.

\subsection{Acceleration factor $\xi$}
The adaptive coefficient has generally been influenced by both first- and second-order moments. However, alternative methods have been proposed to address challenges such as the risk of convergence to local optima and generalization issues. For instance, the Tiger\cite{tigeropt} and Lion\cite{lion} optimizers rely on the sign of the gradient for adaptation, while Adabelief\cite{adabelief} replaces the squared gradient with the difference between the gradient and momentum. In this study, we tackle similar issues by introducing an additional parameter\cite{diffgrad}, $\xi$, which is estimated using the following equation.
\begin{equation}
\xi \leftarrow \frac{1 + e^{-0.5}}{1 + e^{-\lvert x_{t-1} - P \rvert}}
\end{equation}

The parameter $\xi$ is computed based on the final difference between the squared gradient's magnitude and the $P$ factor (the Non-linear Transformation of the dot product of the gradient and momentum). This estimation helps determine whether convergence has been reached and whether to adjust the learning rate accordingly. In the example shown in Fig.\ref{fig:demo}, the gradients exhibit periodic and dramatic changes, which often cause most optimizers to get trapped in a local optimum due to rapidly diminishing gradient values.

However, the term $\lvert x_{t-1} - P \rvert$ enables the process to be dominated by $x_{t-1}$ under normal conditions, while the $P$ factor takes precedence during rapid gradient changes. This mechanism reduces the risk of converging to a local minimum. As depicted in the figure, the Frankenstein adaptive factor demonstrates a recurring pattern, with the $P$ factor dominating $\xi$ during an ``idling phase'' until a significant gradient change occurs.


\begin{algorithm}[tb]
\caption{Frankenstein optimizer}
\label{alg:frank_opt}
\begin{algorithmic}
    \REQUIRE learning rate $\alpha$
    \STATE {\bfseries initialize} $\theta_0, m_0, v_0, t \gets 0, x_0 \gets \alpha, \alpha_0 \gets \alpha$
    \REPEAT
        \STATE $t \gets t + 1$
        \STATE $g_t \gets \nabla_\theta f_t(\theta_{t-1})$
        \STATE $\beta_{1,t} \gets 1 - \text{Clip}(0.1 \sqrt{\alpha_t / 10^{-3}}, 0.05, 0.99)$
        \STATE $P \gets \cos^{-1} (\tanh(m_{t-1} \odot g_t)) / \pi$
        \STATE $x_t \gets g_t^2 + \epsilon$,  $v_t \gets \max(v_{t-1}, x_t)$
        \STATE $\rho \gets \log(\text{Clip}((e^{1} + \sqrt{x_t} + 0.5 - P), e^{0.8}, e^{1.05}))$
        \STATE $\xi \gets (1 + e^{-0.5}) / (1 + e^{-|x_{t-1} - P|})$
        \STATE $m_t \gets \rho \beta_{1,t} m_{t-1} - \alpha_t g_t \xi / \sqrt{v_t}$
        \STATE $\theta_t \gets \theta_{t-1} + \beta_{1,t} m_t - \alpha_t g_t \xi / \sqrt{v_t}$
        \STATE $\beta_{2,t} \gets 1-\frac{x_t}{x_{t-1}} |0.5 - P|$
        \STATE $v_t \gets \beta_{2,t} v_t + \left(1 - \beta_{2,t}\right) x_t$
    \UNTIL{$\theta_t$ converged}
\end{algorithmic}
\end{algorithm}

For a more precise characterization of the optimizer’s behavior, Table \ref{tab:rho_conditions} documents its performance metrics and the dynamics of primary update parameters across successive training stages (Early Training, Mid‑Training, Near Convergence) and in extreme cases (Gradient Spikes; Vanishing Gradients).

\begin{table*}[ht]
\caption{Summary of conditions: behavior of $\rho$ based on gradient square $x_t$, factor $P$ and $\beta_2$.}
\label{tab:rho_conditions}
\vskip 0.15in
\begin{center}
\begin{small}
\begin{tabular}{cccccc}
\toprule
Condition & $x_t$ & $P$ & $\rho\beta_1$  & $\beta_2$ & Optimizer Implication \\
\midrule
Early Training & Large & $\sim  0$ & Large ($\sim 0.95$) &  $\sim 0.5 $ & SGD with high momentum\\
Mid-Training & Moderate & Increasing & Decreasing & Increasing & Adam with low $\beta_2$ \\
Near Convergence & Small & $\sim 0.5$ & $\sim 0.9$ & high $\beta_2$ & Adam with high $\beta_2$ \\
Gradient Spike & Very large & $>0.5$ & $<0.9$ & very low $\beta_2$ & SGD with low momentum \\
Vanishing Gradients & Very Small & $>0.5$ & $\sim 0.9$ & $\approx1$ & Adam with ultra high $\beta_2$ \\
\bottomrule
\end{tabular}
\end{small}
\end{center}
\vskip -0.1in
\end{table*}

\section{Experimental results}


\subsection{Image classification}

\subsubsection{Learning from scratch}

We assessed the performance using the ImageNet-1K\cite{ImageNet} dataset. With the timm\cite{timm} framework, we trained both ResNet18 and ResNet50\cite{resent} models, repeating the training process five times. Given that many studies evaluating optimizers compare ImageNet training over 200 epochs, it is impressive that Frankenstein achieves comparable results to Adam in just 40 epochs, as show in Table \ref{tab:imagenet}. It is also comparable to state-of-the-art optimizers like Adan and AdaFisher in just 80 epochs in ResNet18 model architecture.

\begin{table}[t]
\caption{Validation Top-1 error rate on ImageNet-1K classification. *\cite{adabelief}, $+$\cite{padam},$\oplus$\cite{adan}}
\label{tab:imagenet}
\vskip 0.15in
\begin{center}
\begin{small}
\resizebox{\columnwidth}{!}{%
\begin{tabular}{lll}
\toprule
Optimizer & ResNet18 & ResNet50 \\
\midrule
Adam \cite{adam} & 33.46* & 23.10$\oplus$\\
Lookahead-Adam \cite{lookahead} & - & 24.51\\
Radam \cite{radam} & 32.38* & - \\
AMSGrad \cite{amsgrad} & 32.31$+$ & - \\
Adabound \cite{adabound} & 31.87* & - \\
PAdam \cite{padam} & 29.93$+$ & - \\
AdaBelief \cite{adabelief} & 29.92* & - \\
SGD \cite{sgd} & 29.77* & 23.00$\oplus$\\
LAMB \cite{lamb} & 31.36$\oplus$ & 23.00$\oplus$\\
AdaFisher \cite{adaFisher} & - & 22.99\\
Adan \cite{adan} & 29.10$\oplus$ & \textbf{21.90}$\oplus$\\
\midrule
\textbf{Frankenstein}-40 epoch (Ours) & 31.24 & 25.5\\
\textbf{Frankenstein}-80 epoch (Ours) & 29.13 & 22.9\\
\textbf{Frankenstein}-120 epoch (Ours) & \textbf{28.75} & 22.44\\
\bottomrule
\end{tabular}
}
\end{small}
\end{center}
\vskip -0.1in
\end{table}


\subsubsection{Fine-tuning}
\begin{figure*}[h]
    \centering
    \includegraphics[width=1\linewidth]{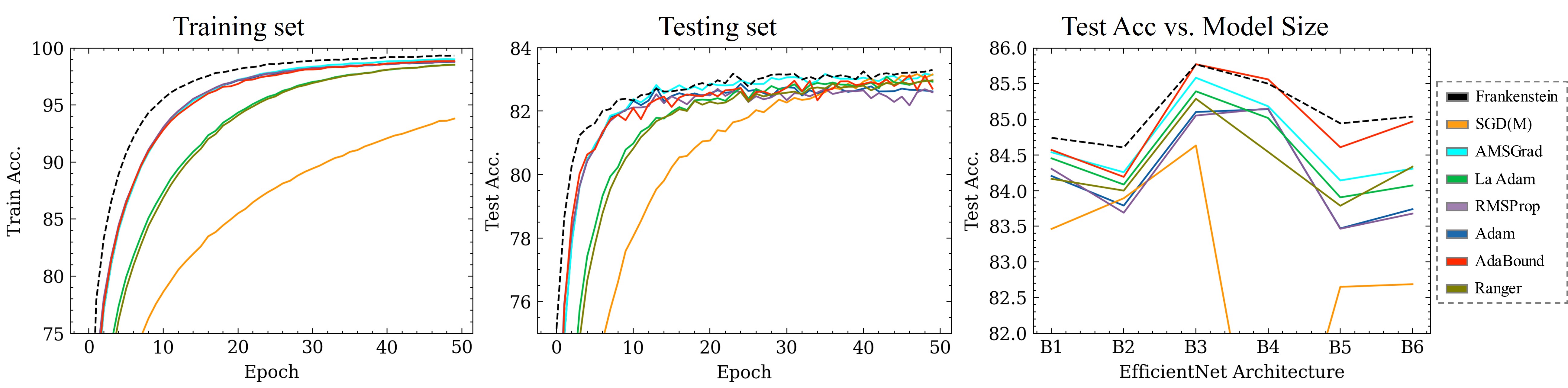}
    \caption{Testing set accuracy profiles of various optimizers on EfficientNetB0 with CIFAR-100 (a) and (b). (c) Testing set accuracy according to architectures (EfficientNetB0-6).}
    \label{fig:cifar10}
\end{figure*}

Transfer learning across different domains has become a widely adopted approach in deep learning. In this study, we trained various optimizers on EfficientNet architectures ranging from B0 to B6, fine-tuning parameters originally learned from ImageNet-1K on the CIFAR-100 dataset. We reconfigured the initial classifier with a MLP layer to align with the target number of classes. Notably, we applied full parameter fine-tuning throughout to achieve optimal performance. Frankenstein got the best accuracy for B1, B2, B5, and B6 while providing the fastest convergence as show in  Fig.\ref{fig:cifar10}.


\subsection{Language modeling}


\subsubsection{Natural language understanding}
The experiment used the IMDB movie review dataset\cite{imdb_dataset} to fine-tune a BERT\cite{bert} model for binary sentiment classification. Two GRU\cite{gru} layers were integrated after the transformer\cite{transform} architecture to perform the final classification.

\begin{figure*}[h]
    \centering
    \includegraphics[width=\linewidth]{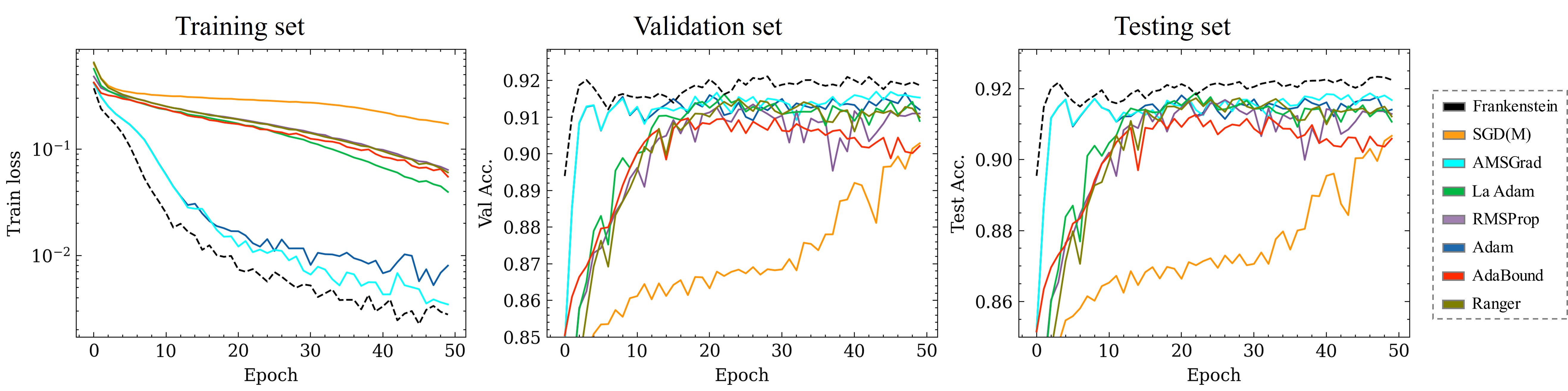}
    \caption{Training, validation and testing set accuracy with different optimizers deal with IMDB movie reviews using BERT. }
    \label{fig:nlp}
\end{figure*}

The dataset was divided into training, validation, and test sets, with 30\% of the data allocated for validation and the remaining 70\% for training. The test set was kept separate and used solely for performance evaluation. All optimizers were tested under consistent hyperparameters to ensure fair comparison. Specifically, a learning rate of $2 \times 10^{-4}$ and a batch size of 128 were applied across all models, with training conducted over 50 epochs. Frankenstein also achieved the fastest convergence among its competitors, as well as superior generalization, as shown in Fig. \ref{fig:nlp}.

\subsubsection{Natural language generation}
\begin{figure}[h]
    \centering
    \includegraphics[width=\linewidth]{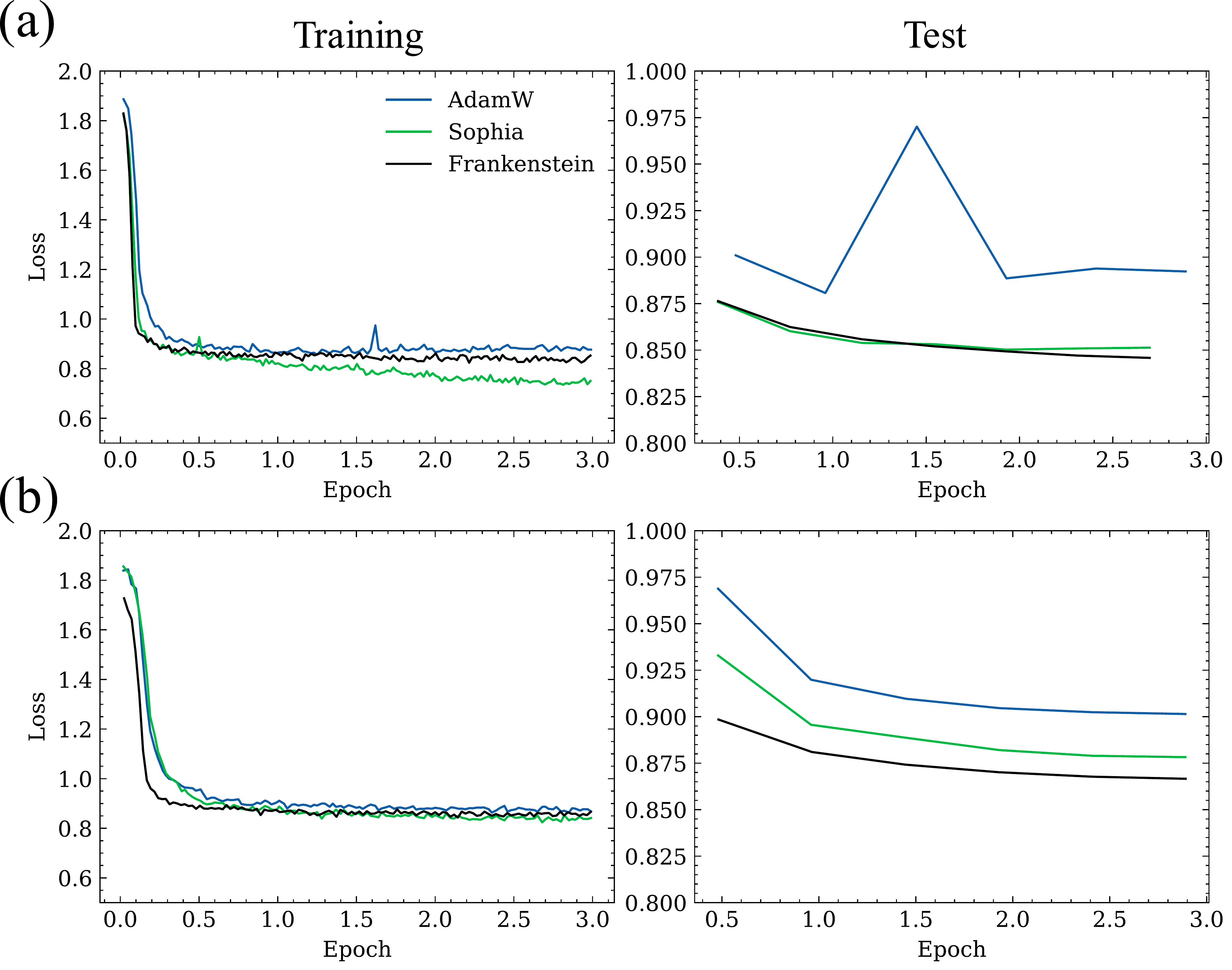}
    \caption{Training and testing losses for text lengths (a) 256 and (b) 512 from the instruction-tuning process of the LLaMA 7B model.}
    \label{fig:peft}
\end{figure}

The instruction fine-tuning of the LLaMA-7B\cite{llama} using Parameter-Efficient Fine-Tuning (PEFT)\cite{peft} will serve as a benchmark test. LoRA\cite{lora} is integrated by adding branches with rank dimension 8 to each layer's \texttt{q\_proj} and \texttt{v\_proj} modules. The Alpaca dataset\cite{alpaca}, generated via self-instruct\cite{self-Instruct} from \texttt{text-davinci-003}, used while 2,000 samples reserved for the test. The model is trained with a batch size of 120 for 3 epochs with a cosine annealing schedule, and the performance is evaluated using maximum token lengths of 256 and 512. As shown in Fig.\ref{fig:peft}, Sophia achieves lower training loss than Frankenstein but worse test loss, indicating overfitting. In addition, the Adam optimizer is affected by grouping sequences based on their length during training, which results in performance spikes on the test set. This behavior reflects an underlying issue of training instability.

\subsection{Few-shot learning}
\begin{figure}[t]
    \centering
    \includegraphics[width=\linewidth]{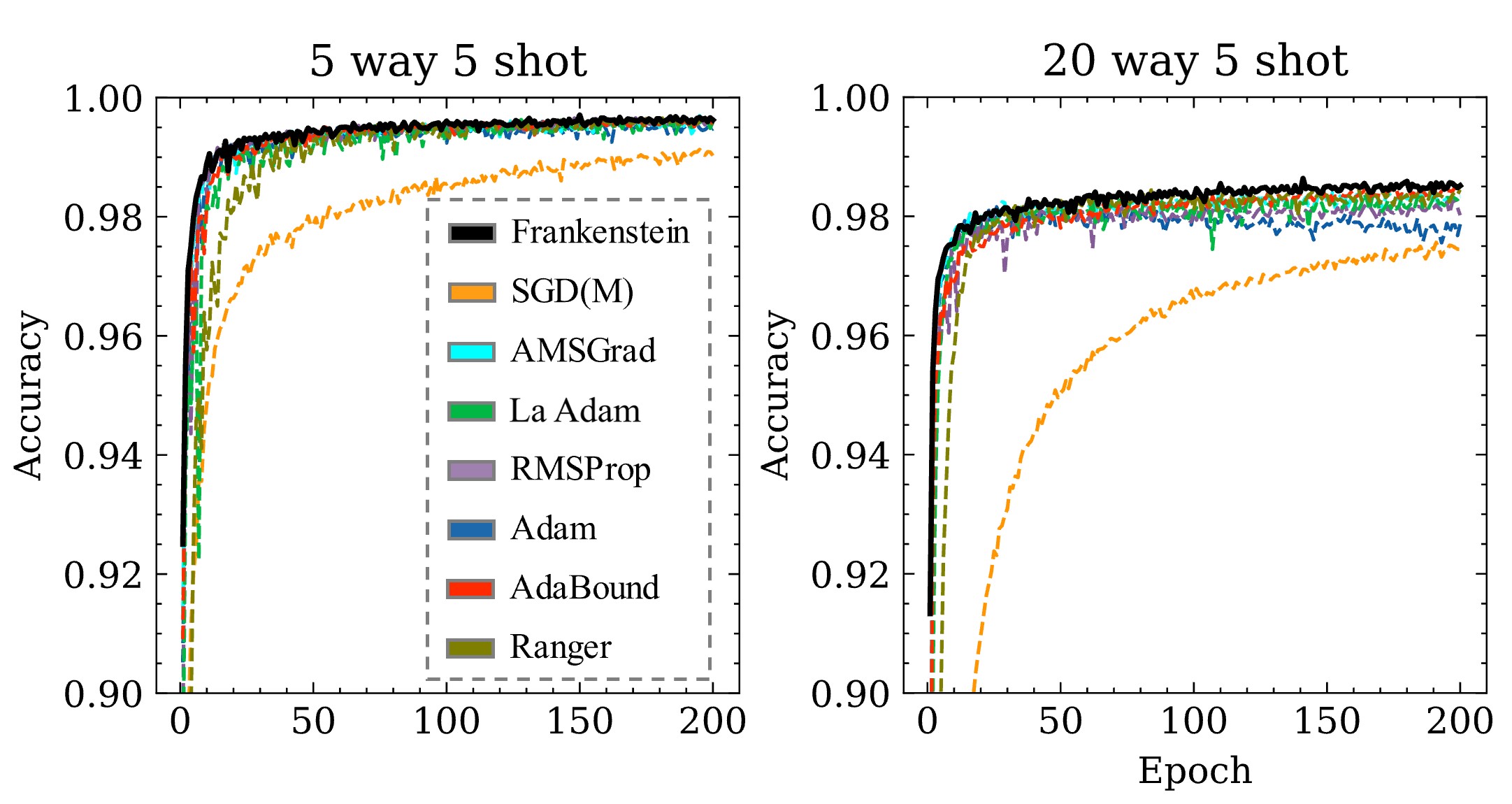}
    \caption{The testing result of several optimizers on the Model-Agnostic Meta-Learning task with multiple strategies.}
    \label{fig:maml}
\end{figure}

This experiment uses MAML (Model-Agnostic Meta-Learning)\cite{maml1,maml2,maml3}, a framework involving a two-stage optimization process, as a performance benchmark. During the few-shot learning process, MAML undergoes meta-training across multiple classification tasks, allowing the model to learn generalizable features and effectively initialize its parameters. MAML can quickly adapt to Omniglot\cite{omniglot} classification tasks with high accuracy, even with only a few update steps and a minimal number of samples. The learning process uses an optimizer in the inner loop to adjust the model based on transferable data, while an optimizer in the outer loop evaluates the model’s overall learning effectiveness. This setup allows us to observe the learning dynamics of different algorithms. For simplicity, gradient descent is used as the update method for the inner loop. The experiments were conducted under 5-way and 20-way settings with 5-shot conditions. Frankenstein exhibits faster convergence and higher accuracy than others (Fig.\ref{fig:maml}).

\subsection{Material modeling and simulations}
Gradient-based optimization plays a crucial role in scientific simulations as well as deep learning, closely linked to simulation accuracy. 

\subsubsection{Energy minimization} 
In energy minimization experiments, it is necessary to minimize both the atomic position to reduce corresponding potential energy and ensure that the model is reasonable at the start of the simulation. This process involves multiple iterations until the energy converges. 
The classic Lennard-Jones potential problem with 38 atoms is demonstrated during the study. The experiment involves randomly initializing 1,000 structures and testing for convergence when the total force drops to the $1 \times 10^{-2}$ level. By analyzing the average, minimum, and maximum number of steps required by common algorithms, the differences between algorithms in molecular dynamics during energy minimization are highlighted. The entire energy minimization process is performed with Atomic Simulation Environment (ASE)\cite{ase} framework, with all initial atomic structures sourced from Lennard-Jones 38 of OptBench\cite{opt_benchmark}. In this benchmark, the number of force calls needed to reach the global minimum is recorded. Frankenstein reaches the global minimum with the least average steps as shown in Table \ref{tab:lj38}. However, the maximum step count is relatively high, possibly due to the different structure of the energy landscape compared to that with flat minima in deep learning.

\begin{table}[h]
\caption{Minimization Benchmarks on Lennard-Jones 38 Clusters}
\label{tab:lj38}
\vskip 0.15in
\begin{center}
\begin{small}
\begin{tabular}{lrrr}
\toprule
Algorithm & $\overline{N}$ & Min $N$ & Max $N$\\
\midrule
Steepest Descent & 4901 & 1355 & 9982 \\
BFGS  & 463 & 243 & 8210 \\
Conjugate Gradient  & 453 & 207 & 1153 \\
Fire  & 656 & 208 & \textbf{1000} \\
Frankenstein  & \textbf{383} & \textbf{183} & 3469 \\
\bottomrule
\end{tabular}
\end{small}
\end{center}
\vskip -0.1in
\end{table}

\begin{figure}[h]
    \centering
    \includegraphics[width=\linewidth]{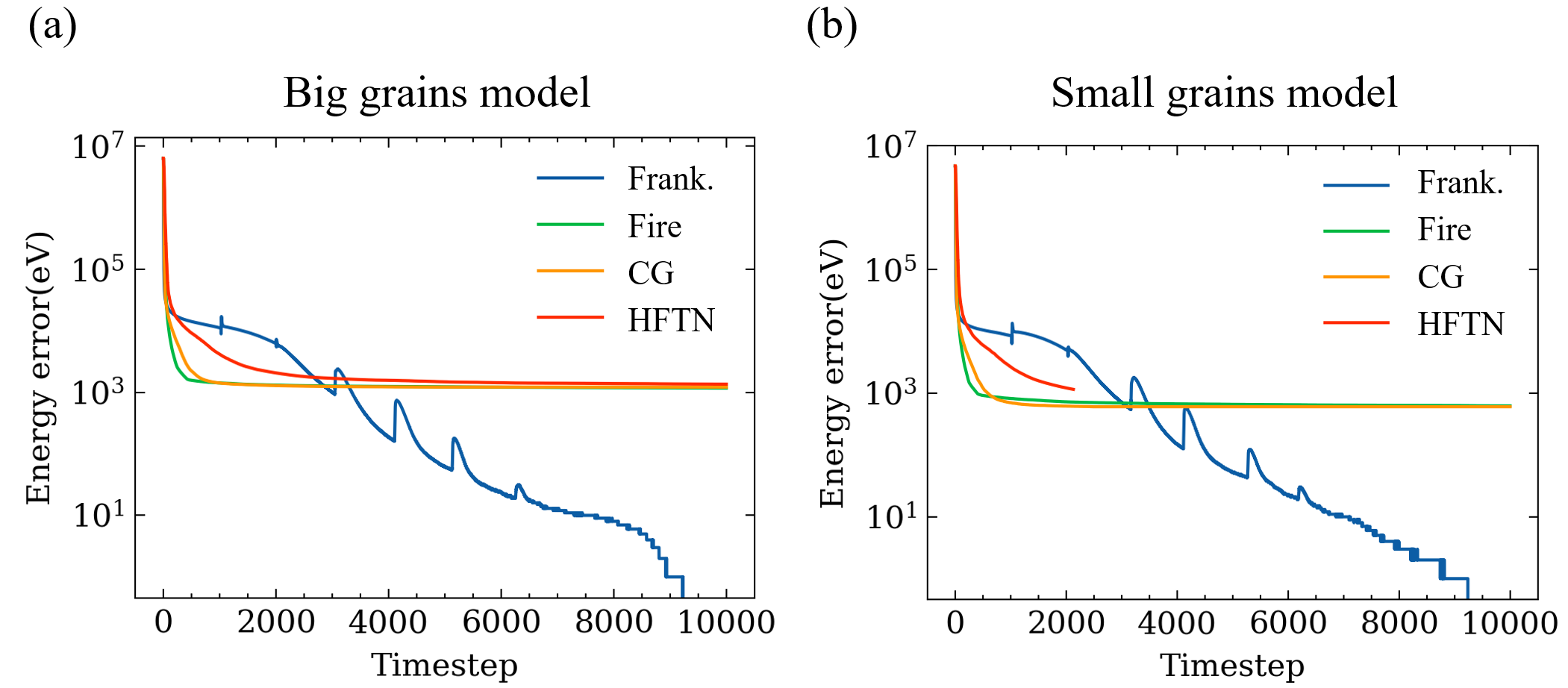}
    \caption{The optimization benchmark for the energy minimization algorithm applied to polycrystal high-entropy alloys according to the grain size. The values indicate the difference between the potential energy of the entire system and the best result found to date (i.e., the result achieved by the Frankenstein search).}
    \label{fig:md_min}
\end{figure}

To investigate the case of more complex structures, a polycrystalline high-entropy alloy (HEA) model composed of Co, Ni, Cr, Fe, and Mn is generated using Atomsk\cite{Atomsk}. The energy minimization of the entire system uses the Second Nearest-Neighbor Modified Embedded-Atom Method (2NN MEAM)\cite{2nn_meam,hea_pe} potential. The system consists of 5.5 million atoms forming a periodic box with 40 × 40 × 40 nm dimensions. All optimization processes are performed using LAMMPS\cite{LAMMPS}, and the minimization methods are executed according to recommended parameters. Three commonly used optimization methods are adopted as references: FIRE\cite{fire_lammps}, conjugate gradient (CG), and the Hessian-free truncated Newton algorithm(HFTN). Notably, the best-observed value so far is considered the minimum due to the absence of a well-defined global minimum energy value. Compared to widely used competitors, Frankenstein effectively escapes local optima and reaches the lowest energy state (Fig.\ref{fig:md_min}).

Additionally, the system's microstructural analysis after energy minimization is performed using dislocation analysis\cite{dxa} (Table \ref{tab:mini_poly}). The atomic model optimized by our algorithm approaches the accurate minimum potential energy and achieves a significant reduction in the total length and the number of dislocations. This enhanced stability of our Frankenstein suggests that future experiments will be less prone to biases arising from initialization-related defects.

\begin{table}[t]
\caption{Energy minimization on polycrystal high entropy alloy}
\label{tab:mini_poly}
\vskip 0.15in
\centering
\begin{small}
\begin{tabular}{lrrr}
\toprule
Method & Error PE & Dislocation & Dislocation\\
 & (eV)$\downarrow$ & segments$\downarrow$ & length (nm)$\downarrow$\\
\midrule
Frankenstein & \textbf{0} & \textbf{8406} & \textbf{7775} \\
Fire & 1172 & 8627 & \ 8369 \\
CG & 1216 & 9469 & \ 9328 \\
HFTN & 1358 & 8853 & \ 8426 \\
\bottomrule
\end{tabular}
\end{small}
\vskip -0.1in
\end{table}

\subsubsection{Collaboration with materials discovery}

Matbench Discovery\cite{matbench} is introduced as a framework to support various NN potentials for assessing the accuracy of predictions of solid-state thermodynamic stability. MACE\cite{mace}, as a pre-trained foundation model for potentials, is incorporated and benchmarked alongside various energy minimization methods. During the process, models are capped at a maximum of 500 iterations or a force threshold below $1 \times 10^{-2}$ as the stopping criterion. The experiment applies relaxation to the initial structures of 257k inorganic crystals from the WBM dataset\cite{wbm}. Two key indicators are used to evaluate the optimization methods: the number of systems with an energy error $> 1$ eV/atom (indicating convergence failure) and the average energy error after filtering. Frankenstein fails the least and, on average, reaches a lower energy state as shown in Table \ref{tab:mace}.

\begin{table}[t]
\caption{Matbench Discovery Benchmarks on MACE}
\label{tab:mace}
\vskip 0.15in
\begin{center}
\begin{small}
\begin{tabular}{lccc}
\toprule
 & LBFGS & FIRE & Frankenstein\\
\midrule
Error $\textgreater$ 1ev/atom (count) & 1229 & 1076 & \textbf{1034} \\
MAE (meV)  & 97.64 & 96.3 & \textbf{95.64} \\
\bottomrule
\end{tabular}
\end{small}
\end{center}
\vskip -0.1in
\end{table}

\subsubsection{Neural network quantum state}
\begin{figure}[h]
    \centering
    \includegraphics[width=\linewidth]{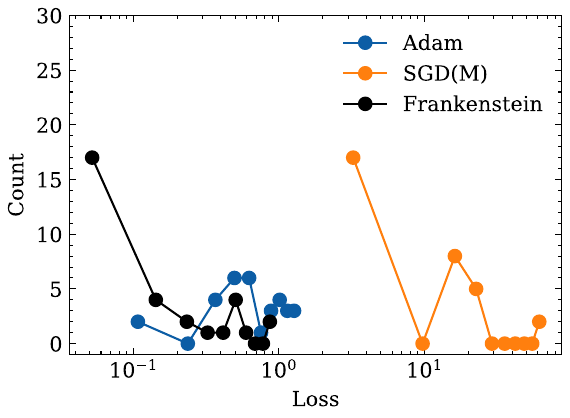}
    \caption{Distribution of loss after 10k iterations for different optimizers over 32 tests of simulating a quantum system with neural network training to approximate the lowest energy eigenstate of a Hamiltonian.}
    \label{fig:qnn}
\end{figure}

In this experiment, we construct a 2D quantum spin lattice model $H_{J1J2}$ with nearest-neighbor (\( J_1 \)) and next-nearest-neighbor (\( J_2 \)) interactions, where the coupling constants are set to fully frustrated\cite{choo2019two} as \( J_1 = 1 \) and \( J_2 = 0.5 \), which lately to indicate to have numerical sign problem for the neural network quantum state\cite{szabo2020neural}. The Hamiltonian is generated using the Pauli representation and converted into a sparse matrix format for computational efficiency. We then calculate the eigenvalues using the neural network quantum state $\ket{V_{nn}}$ to estimate the ground state energy $\frac{\bra{V_{nn}}H_{J1J2}\ket{V_{nn}}}{\braket{V_{nn}}{V_{nn}}}$, providing an indicator of the system’s stability. A NN, composed of layered linear and ReLU activations, generates a normalized complex parameter vector to optimize the initial quantum states. Each optimizer undergoes 32 convergence tests to ensure stability and consistency. The model's performance is evaluated by minimizing energy discrepancies, offering insights into optimizing quantum states for hybrid quantum-classical computing\cite{zhang2022variational} and materials simulations\cite{yoshioka2021solving}. The landscape of NNQS is notorious to train, but Frankenstein achieves the lowest loss frequently, as shown in Fig.\ref{fig:qnn}.

\section{Analysis}

We examined the changes in NN weights and representations during training, focusing on their impact on generalization and the learning process\cite{losslandscape}. To visualize the weights of the entire network (EfficientNetB0, which contains 5.3 million parameters) in a 2D space, the weights at each training epoch were recorded and reduced to two dimensions using PCA. The produced network-specific hyperplane encapsulates the optimizer's training trajectory. Based on the PCA1 and PCA2 coordinates of each point in this space, the model's weights were reconstructed, and performance was evaluated on the CIFAR-100 testing set.

\begin{figure}[h]
    \centering
    \includegraphics[width=\linewidth]{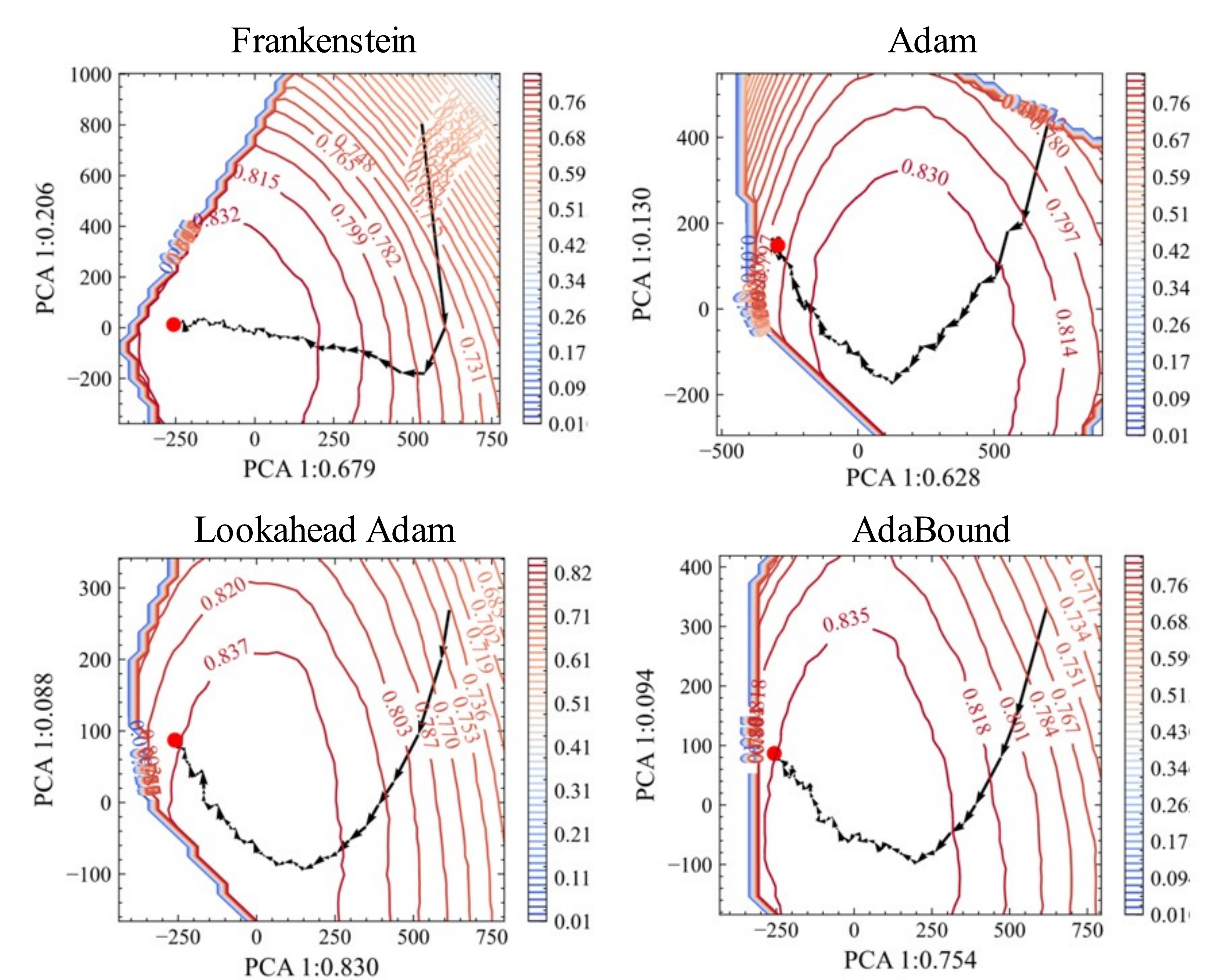}
    \caption{The learning process of different optimizers during transfer learning on EfficientNetB0 using the CIFAR-100 dataset. The contour lines represent the model's accuracy on the test set.}
    \label{fig:loss_landscape}
\end{figure}

The training trajectory is illustrated on the plane using black arrows. This hyperplane reveals distinct preferences for different optimizers, as shown in Fig.\ref{fig:loss_landscape}. For instance, Adam tends to overfit, as it quickly deviates from the flat minima region. In contrast, our Frankenstein exhibits a stabilized trajectory in a landscape with superior generalization after reaching the target region. 



\begin{figure}[h]
    \centering
    \includegraphics[width=0.9\linewidth]{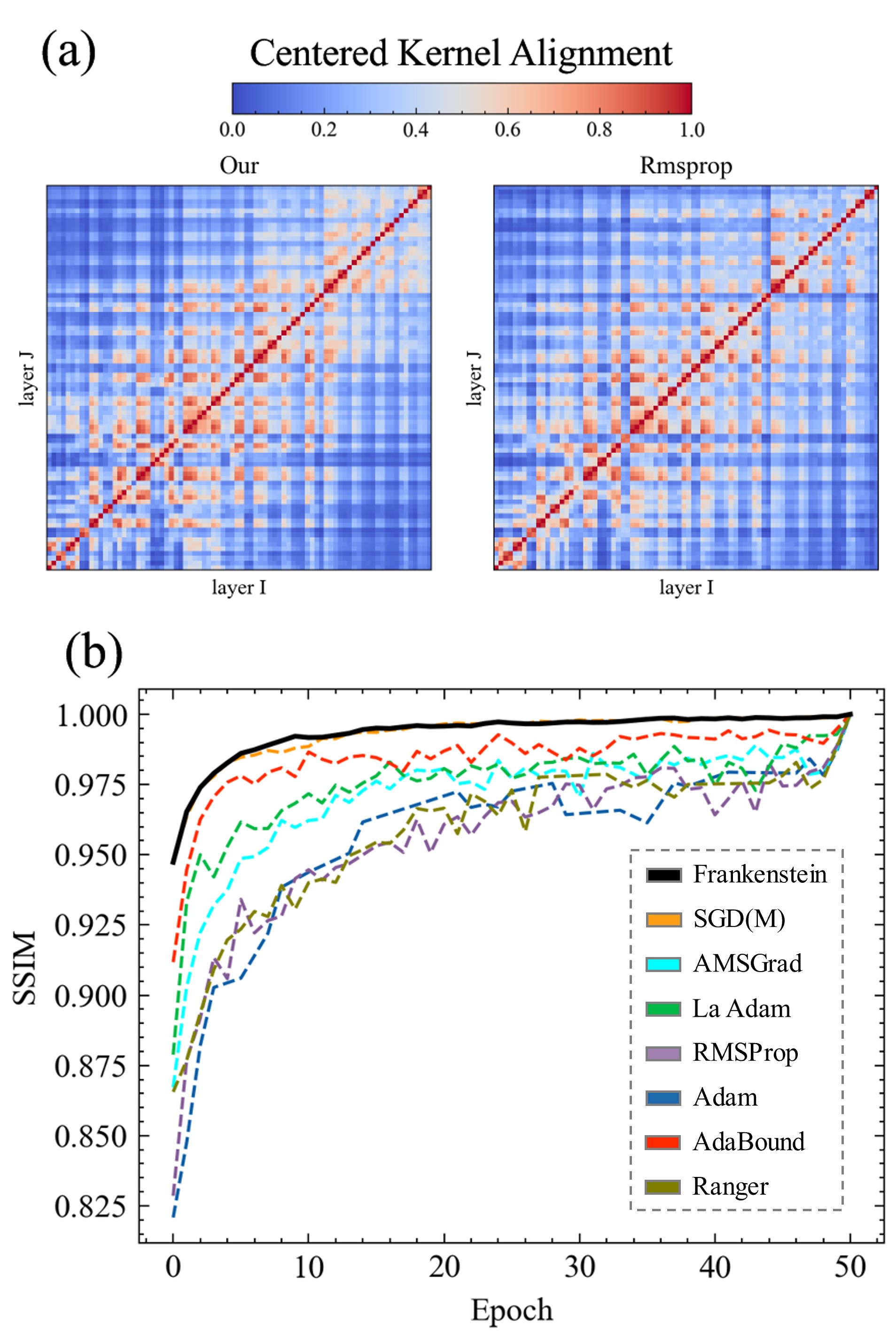}
    \caption{(a) Centered kernel alignment (CKA) show the difference of representations between RmsProp and Frankenstein. (b) History profiles of structural similarity index between different epoch and 50th epoch CKA matrix, to define the effective forming period of the functional network.}
    \label{fig:cka}
\end{figure}


In addition to analyzing NN weights, examining changes in NN representations is a valuable approach to understanding how low-level features propagate through layers and transform into high-level semantics. The Centered Kernel Alignment (CKA) method\cite{cka,cka2,improving_representations}
 has been proposed as an effective tool for addressing such problems, as it highlights representation changes across different architectures. By approximating an identity matrix, where the information between layers becomes maximally decorrelated, each layer sufficiently transforms the information and reveals the influence of architecture on model representations.

In this study, we present the results of CKA analysis in Fig. \ref{fig:cka}(a), comparing the performance of RMSprop with our proposed algorithm. The results clearly show that our method effectively prevents low-level features from leaking into layers near the output. Addit in fine-tuned models, a cross-optimizer analysis is shown in Fig. \ref{fig:cka}(b), where CKA maps from each epoch are compared to the final converged result. Matrix similarity is measured using the Structural Similarity Index (SSIM) metric. It is important to note that while task performance metrics such as loss and accuracy can be clearly quantified, they do not adequately capture whether the learned representations undergo substantial changes during fine-tuning. This aspect is crucial for understanding the model’s zero-shot capability. To investigate this, we use representations pre-trained on large-scale datasets, such as ImageNet, as a strong baseline and then apply simple fine-tuning on a downstream task, such as CIFAR-100. If the model weights remain largely unchanged while still achieving strong performance on the downstream task, this suggests that the representations are stable. This stability can be assessed by measuring the CKA similarity between the final model and the initial pre-trained model. A higher similarity indicates minimal representational drift. In our experiments, we observed that models more prone to overfitting, such as those trained with Adam or RMSProp, showed greater divergence in their final representations

\section{Conclusion}
In this paper, we proposed the Frankenstein optimizer, dynamically adjusting its momentum coefficients to improve convergence. We comprehensively evaluated our Frankenstein to demonstrate its versatility and effectiveness across broad applications. It offers superiority in complex deep learning tasks such as image classification, instruction fine-tuning, sentiment classification, and collaborative optimization for meta-learning. It also has potential to advance materials design by treating challenges such as structural optimization, energy minimization, and defect reduction. In quantum applications, our optimizer efficiently navigated the `barren plateau' while others didn't. Additionally, our analyses using CKA and the loss landscape visualization provided deeper insights into the behavior of adaptive optimizers during training. Overall, the Frankenstein optimizer has demonstrated its ability to tackle impactful problems in AI, science, and engineering, paving the way for breakthroughs across multiple domains.


\section*{Impact Statement}
Recently, many advances in science and engineering have increasingly relied on machine learning methods to tackle existing challenges. Our algorithm, applicable for any gradient-based trainable model, is designed to improve computational efficiency, robust generalization, and adaptability across diverse tasks. Our research has the potential to be applied to various technologies, including the superintelligence, material and drug design, and quantum security. However, we believe that it is not necessary to specifically discuss their social and ethical implications at this stage.

This study also emphasizes the recombination of various findings, showcasing how rational integration of multiple prior improvements can enhance performance. The CKA and landscape methods serve as valuable tools for analyzing learning dynamics, providing insights into the behavior of both optimizers and models. During this process, we observed that certain popular algorithms enjoy a privileged \cite{privileged_adam} status across different fields.
Furthermore, the values of our momentum coefficients, $\beta$ and $\beta_2$, remain within the range of 0 to 1. As a result, we did not perform a formal convergence analysis.

\section*{Acknowledgments}
This work was supported by SES AI. Computational resources were partially provided by the Texas Advanced Computing Center (TACC) and the MIT SuperCloud. Additional support was provided by the “High Entropy Materials Center,” funded through the Featured Areas Research Center Program under the Higher Education Sprout Project by the Ministry of Education (MOE), Taiwan.

\clearpage

\bibliography{reference}
\bibliographystyle{icml2025}

\newpage
\appendix
\onecolumn

\section{More experiment detail}
\begin{figure}[h]
    \centering
    \includegraphics[width=0.4\linewidth]{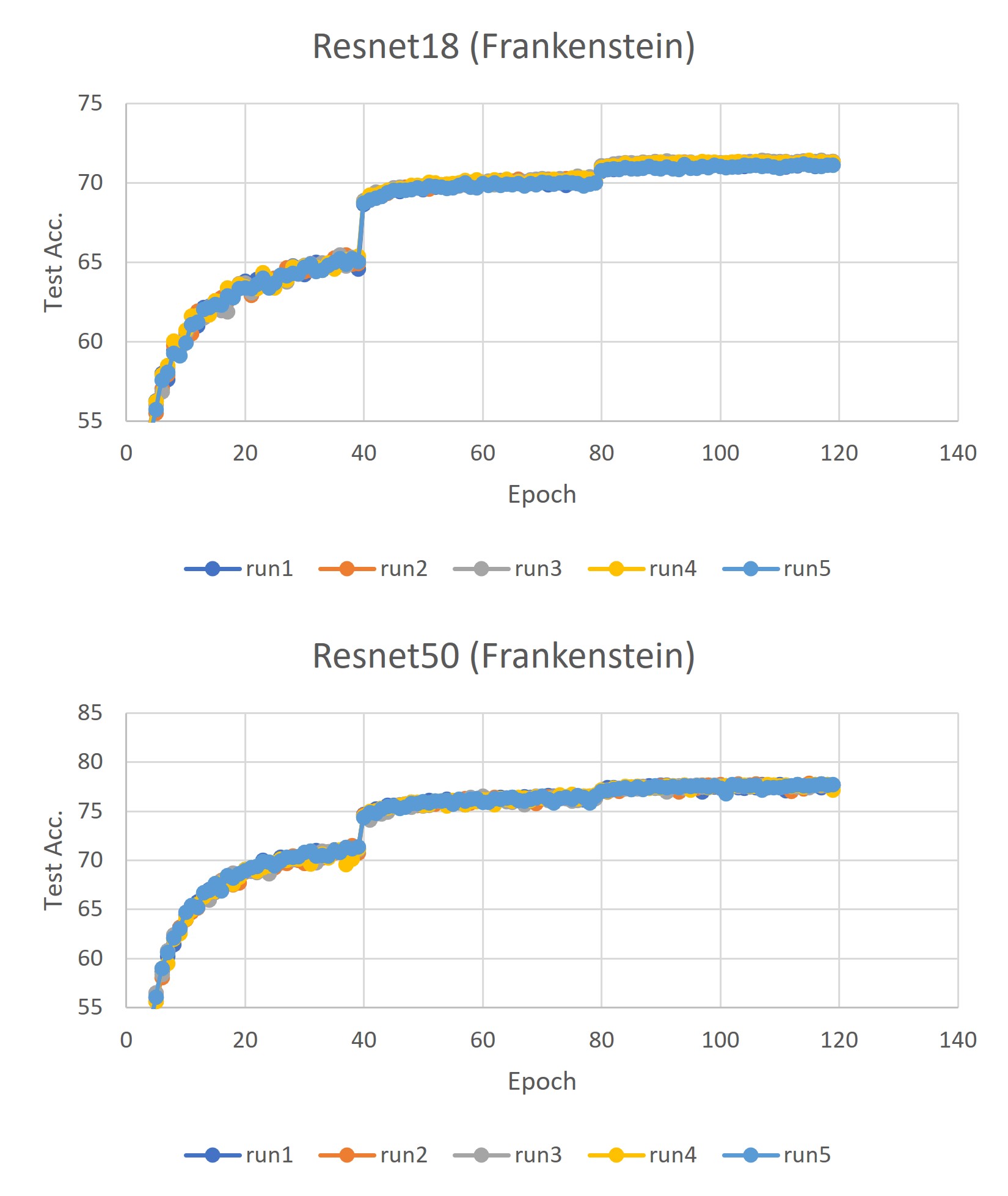}
    \caption{Test accuracy progression during training for Resnet18 and Resnet50 Frankenstein optimizer. The curves represent multiple independent runs, illustrating training stability and convergence.}
    \label{fig:imagenet_detail}
\end{figure}

\begin{figure}[h]
    \centering
    \includegraphics[width=0.4\linewidth]{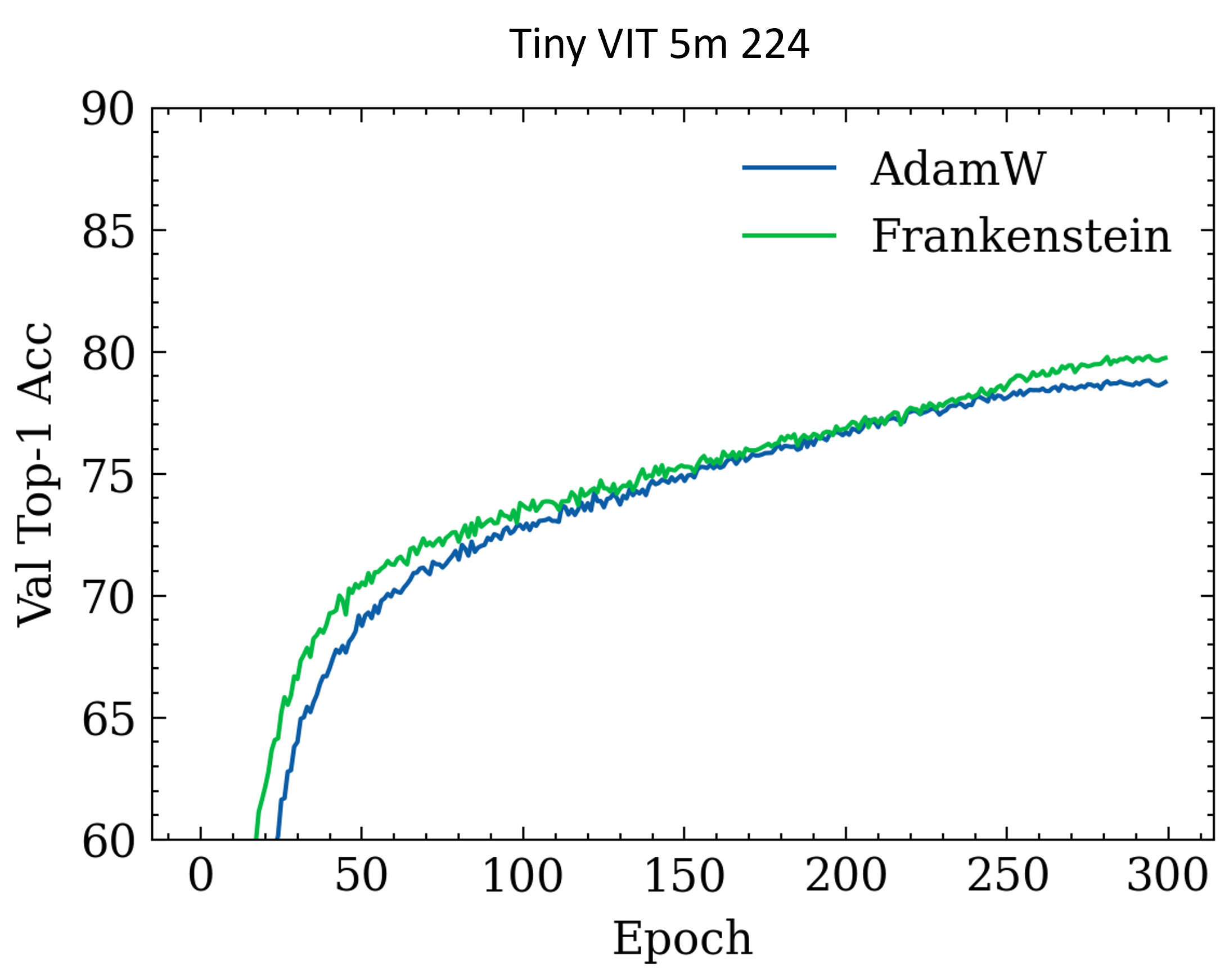}
    \caption{Validation Top-1 Accuracy of Tiny ViT 5m 224 on ImageNet. Comparing performance with AdamW and Frankenstein optimizers.}
    \label{fig:tiny_vit}
\end{figure}

Regarding the from-scratch training results on ImageNet, as shown in the figure below, we ran experiments using ResNet-18 and ResNet-50 for 120 epochs. Each experiment was repeated 5 times, and only the average values are reported in the fig \ref{fig:imagenet_detail}. The hyperparameters were adopted from the recommended settings in the TIMM library\cite{timm}. For example, for ResNet-18, we used: \texttt{--batch-size 256 (across 4 GPUs), --weight-decay 1e-2, --sched step, --lr 1e-3, --epochs 120, --min-lr 1e-5, --decay-epochs 40, --amp}. No specific hyperparameter search was performed, and the batch size was limited by the GTX 2080Ti 11\,GB VRAM. Additionally, we examined the training behavior of Tiny ViT\cite{wu2022tinyvit} with various optimizers, as shown in Fig. \ref{fig:tiny_vit}.

\clearpage

\section{Hyperparameter}
In our experiments, we conducted a comprehensive evaluation of various optimization algorithms, ranging from classical momentum-based methods to modern adaptive and state-of-the-art optimizers. We assessed their performance across standard tasks like image classification (on CIFAR-10/100) and natural language processing (using BERT fine-tuning), as well as on challenging scenarios such as few-shot learning. To gain deeper insights into optimizer behavior, we employed techniques including the visualization of the loss landscape as a probe and the analysis of neural network representation similarity.

To address concerns regarding experimental reproducibility and provide robust results, all experiments were conducted with five independent runs. The final performance metrics presented are the average results across these repeated trials to ensure reliability.

While adaptive optimizers generally offer greater robustness to hyperparameter choices, we performed systematic tuning for all optimizers on each task to ensure fair comparisons and report results based on their best achievable performance.

For the CIFAR-10/100 and NLU tasks, we conducted a hyperparameter search to identify the optimal settings for each optimizer. Tables~\ref{tab:nlu_strategy} and \ref{tab:cifar_strategy} detail the learning rate search spaces explored. Other hyper-parameters, such as weight decay, were kept consistent across optimizers where applicable, or tuned individually when necessary for optimal performance.

\begin{table}[ht]
\caption{Hyper-parameters searching for NLU tasks.}
\label{tab:nlu_strategy}
\vskip 0.15in
\begin{center}
\begin{small}
\begin{tabular}{cc}
\toprule
Learning rate & Batch size \\
\midrule
$5 \times 10^{-4}$ & 128 \\
$1 \times 10^{-4}$ &       \\
$5 \times 10^{-5}$ &       \\
If using SGD, lr$\times$10 & \\
\bottomrule
\end{tabular}
\end{small}
\end{center}
\vskip -0.1in
\end{table}

\begin{table}[ht]
\caption{Hyper-parameters searching for Cifar tasks.}
\label{tab:cifar_strategy}
\vskip 0.15in
\begin{center}
\begin{small}
\begin{tabular}{cc}
\toprule
Learning rate & Batch size \\
\midrule
$5 \times 10^{-3}$ & 128 \\
$1 \times 10^{-3}$ & 512    \\
$5 \times 10^{-4}$ &       \\
$1 \times 10^{-4}$ &       \\
If using SGD, lr$\times$10 & \\
\bottomrule
\end{tabular}
\end{small}
\end{center}
\vskip -0.1in
\end{table}



\clearpage

\section{Update rule comparison of various optimizers}\label{sec:existing_optimizers}
Several adaptations have been developed for various optimizers, each addressing specific characteristics. These improvements include:
\begin{itemize}
\item  Adding momentum and implementing a bias correction mechanism (Adam\cite{adam}).
\item  Resolving EMA's non-convergence issues by adopting a more rigorous approach (AMSGrad\cite{amsgrad}).
\item  Introducing constraints on the adaptive range to enhance stability (Adabound\cite{adabound}).
\item  Allowing for control over adaptation across different powers, enabling adjustable adaptive rates (Padam\cite{padam}).
\item  Using the difference between the current gradient and momentum as an adaptive component(Adabelief\cite{adabelief}).
\end{itemize}

\begin{table}[ht]
\caption{Update rules of various optimizers}
\label{tab:update}
\vskip 0.15in
\begin{center}
\begin{small}
\begin{tabular}{ll}
\toprule
Algorithm & Update Rule \\
\midrule
RMSProp\cite{rmsporp} & $v_t \leftarrow \beta_2 v_{t-1} + (1-\beta_2) g_t^2$ \\
Adam\cite{adam} & $v_t \leftarrow \frac{\beta_2 v_{t-1} + (1-\beta_2) g_t^2}{1-\beta_2^t}$ \\
AmsGrad\cite{amsgrad} & $v_t \leftarrow \max\left(\frac{\beta_2 v_{t-1} + (1-\beta_2) g_t^2}{1-\beta_2^t}, v^{\max}\right)$ \\
Adabound\cite{adabound} & $v_t \leftarrow \text{Clip}\left(\beta_2 v_{t-1} + (1-\beta_2) g_t^2, v_{\min}, v_{\max}\right)t$ \\
Padam\cite{padam} & $v_t \leftarrow \max\left(\frac{\beta_2 v_{t-1} + (1-\beta_2) g_t^2}{1-\beta_2^t}, v_{t-1}\right)^{2p}$ \\
Adabelief\cite{adabelief} & $v_t \leftarrow \frac{\beta v_{t-1} + (1-\beta_2) (g_t - m_t)^2}{1-\beta^t}$ \\
Frankenstein (Ours) & $v_t \leftarrow \beta_2 v_{t-1} + (1 - \beta_2) g_t^2$ ,\\
& \text{ where } $\beta_2 = 1 - \frac{g_t^2}{g_{t-1}^2} \left| 0.5 - P \right|$\\
& \text{ where } $P \gets \cos^{-1} (\tanh(m_{t-1} \odot g_t)) / \pi$\\
\bottomrule
\end{tabular}
\end{small}
\end{center}
\vskip -0.1in
\end{table}


\clearpage

\section{Ablation experiments}
To demonstrate the role of each component in the optimizer, we conducted ablation experiments on an image classification task. Specifically, we trained the EfficientNet-B0 model (initialized randomly) on the CIFAR-10 dataset and used the final testing accuracy as a metric to evaluate the impact on performance. The training process utilized a batch size of 128 and a learning rate of $1 \times 10^{-3}$ for 100 epochs, reducing the learning rate by a factor of 10 every 40 epochs.

The experiments involved modifications to the momentum settings (e.g., decoupling $\beta$ from the learning rate and fixing $\beta=0.9$) and second-order momentum terms (e.g., removing $v_t$, fixing $\beta_2=0.999$, excluding $\max(v_{t-1}, x_t)$, and omitting the EMA of $v_t$). In total, six variations were analyzed to assess their effect on performance using the test set.

\begin{table}[ht]
\caption{Ablation experiments:The ablation study was tested on the CIFAR-10 dataset, with the neural network architecture being EfficientNetB0. Results are the average of 5 experimental runs.}
\label{tab:ablation}
\vskip 0.15in
\begin{center}
\begin{small}
\begin{tabular}{cc}
\toprule
Case & Test set Accuracy\\
\midrule
Full & $93.17 \pm 0.17$ \\
Fix $\beta_2 =$0.999 & $92.32 \pm 0.29$({\textcolor{red}{-0.85}}) \\
Decouple $\beta$ with LR & $92.22 \pm 0.22$({\textcolor{red}{-0.95}}) \\
Adam & $92.12 \pm 0.17$({\textcolor{red}{-1.05}}) \\
Fix $\beta =$ 0.9 & $92.08 \pm 0.28$({\textcolor{red}{-1.09}}) \\
w/o $\max(v_{t-1}, x_t)$  & $91.77 \pm 0.87$({\textcolor{red}{-1.4}}) \\
w/o EMA of $v_t$  & $91.74 \pm 0.42$({\textcolor{red}{-1.43}}) \\
\bottomrule
\end{tabular}
\end{small}
\end{center}
\vskip -0.1in
\end{table}

As shown in the table \ref{tab:ablation}, the performance of each variation is compared with the original algorithm. Notably, when the dynamically adjusted first-order and second-order momentum coefficients ($\beta$ and $\beta_2$) are fixed to constant values—similar to Adam optimizer default parameters—the convergence results are almost equivalent to Adam optimizer final performance.

\clearpage

\section{Adaptive overfit test}
To evaluate whether these adaptive hypotheses introduce additional biases (e.g., bias correction from Adam\cite{adam}, Radam's variance control policy\cite{radam}) that could lead to overfitting, we introduce a simple binary classification task\cite{simple_test}. The inputs and classifications for this task are defined by the following equation:

for \(i = 1, \dots, n\), where \(n = 6\).
\[
x_{ij} = 
\begin{cases} 
y_i & j = 1, \\
1 & j = 2, 3, \\
1 & j = 4 + 5(i - 1), \dots, 4 + 5(i - 1) + 2(1 - y_i), \\
0 & \text{otherwise},
\end{cases}
\]

\begin{figure}[h]
    \centering
    \includegraphics[width=0.5\linewidth]{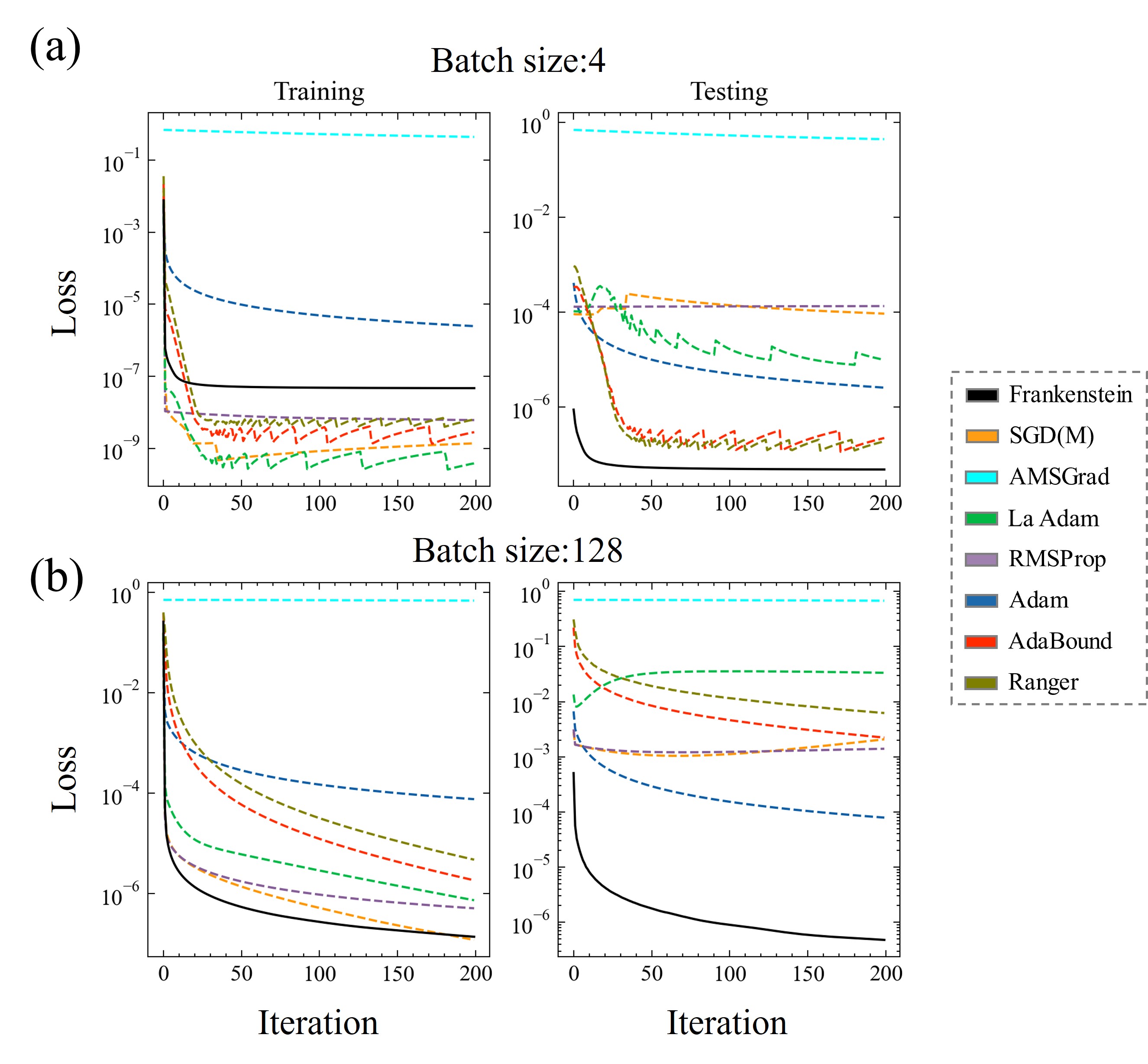}
    \caption{Demonstrate the overfitting behavior of various adaptive algorithms when applied to simple problems, with batch sizes of 4 and 128 shown in (a) and (b), respectively.}
    \label{fig:simple_test}
\end{figure}

In this experiment, we tested two different batch sizes (4 and 128) to explore the relationship between adaptivity and noise\cite{Margin1}, as illustrated in Fig. \ref{fig:simple_test}. Nearly all optimizers achieved extremely low training loss (\(10^{-9}\)), which was limited by precision constraints. Notably, AMSGrad, due to its strategy of maximizing \(v_t\) at each step, struggled to adapt to this type of problem. On the test set (depicted on the right side of the figure), our proposed algorithm demonstrated robust convergence across various conditions, particularly when compared to other adaptive optimizers.

\clearpage

\section{Loss landscape for energy minimization}
\begin{figure}[ht]
    \centering
    \includegraphics[width=\linewidth]{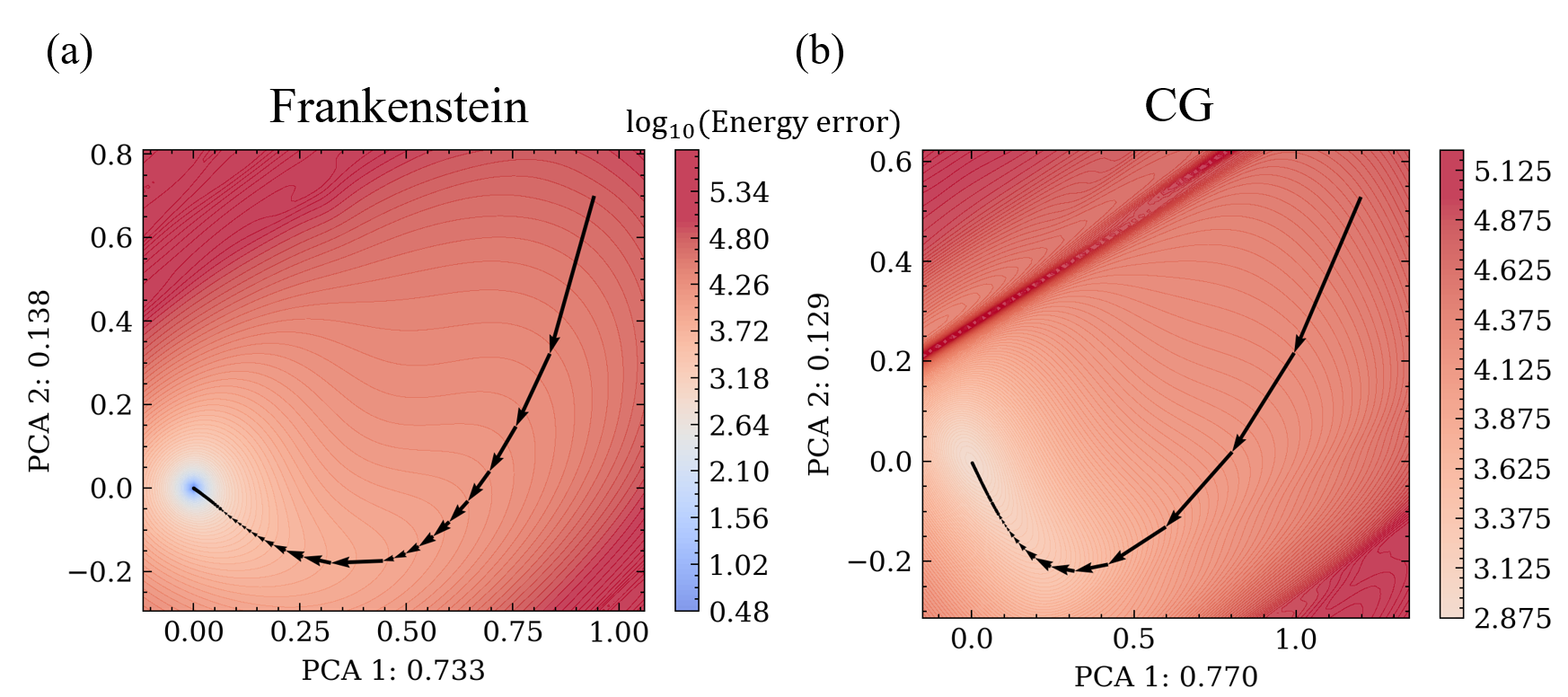}
    \caption{Illustrate the energy minimization process for (a) Frankenstein and (b) CG on polycrystalline high-entropy alloys. The contour lines indicate the difference between the potential energy of the entire system and the best result obtained thus far.}
    \label{fig:md_landscape}
\end{figure}

The loss landscape for visualizing high-dimensional spaces has also been applied to the analysis of the energy minimization process in polycrystalline high-entropy alloys using molecular dynamics models. We compared the behavior of our algorithm with the Conjugate Gradient method on the energy landscape, as illustrated in Fig. \ref{fig:md_landscape}. The acceleration effect achieved by our adaptive approach is clearly demonstrated in the contour plots and learning trajectories, showcasing an optimization process that follows the steepest and most efficient path.

\clearpage
\section{Optimizer setting}
SGD(M): 
Whether the momentum parameter is case sensitive, we implement the most commonly applied value of 0.9. Additionally, the momentum behavior was fellow the Nesterov Accelerated Gradient (NAG), which has more solid theoretical converge guarantees.

RmsProp\cite{rmsporp}:
 In the experiment, the adaptive learning optimizer would take into account being a competitor. RmsProp adjusts the learning rate of each weight with the magnitudes of gradients. Following TensorFlow instruction, we set a smoothing constant with a default value of 0.9.

Adam\cite{adam} \& Amsgrad\cite{amsgrad}: Adam is the most commonly used optimizer, which interjects the bias correction mechanism that significantly helps the initial few training steps. Furthermore, Adam's variant, i.e., AMSGrad, facilitates the problem that can’t converge. We set the $\beta_1$ and $\beta_2$ as 0.9,0.999, respectively.

Lookahead\cite{lookahead}:
Lookahead optimizer is independent of these previous optimizers, which be aid of optimizer. In the experiment, it supports Adam optimizer that parameter set as we mention above. Then, the Lookahead mechanism would manipulate as five sync periods and having 0.5 slow step size.

Adabound\cite{adabound}:
Adabound is a variant of Adam, that dynamic bounds on elements learning rate. Therefore, it enables the training processing with Adam-like initial and SGD-like eventually. We apply the default setting on experiments, i.e., the final learning ratio is 0.1.

Ranger\cite{Ranger}: Ranger is the combination of Lookahead and Rectified Adam. Note that it is regarded as the state-of-the-art optimizer, which can achieve the best performance on a worldwide task. As an influential competitor, we set the suggestion hyperparameter by TensorFlow\cite{tf}.
\end{document}